\documentclass[12pt]{article}
\usepackage{caption,comment,graphicx,hyperref,setspace,subcaption,url}
\usepackage[T1]{fontenc}
\usepackage[margin=1in]{geometry}

\newcommand{\HRule}{\rule{\linewidth}{0.5mm}}

\newenvironment{boxed}
 {\begin{center}
 \begin{tabular}{|p{0.9\textwidth}|}
 \hline
 }
 { 
 \\\hline
 \end{tabular} 
 \end{center}
 }

\makeatletter % since there's an at-sign (@) in the command name
\renewcommand{\@maketitle}{%
 \parindent=0pt % don't indent paragraphs in the title block
 \centering
 {\Large \bfseries\textsc{\@title}}
 {\large \@author}
 \HRule\par %
 \textbf{\underline{Replicated by}:} Shibamouli Lahiri (University of Michigan; Winter 2015)\\\href{mailto:lahiri@umich.edu}{lahiri@umich.edu}
 \HRule\par
}
\makeatother % resets the meaning of the at-sign (@)

\title{Replication of the Keyword Extraction part\\of the paper\\``Without the Clutter of Unimportant Words'':\\Descriptive Keyphrases for Text Visualization}
\author{Jason Chuang and Christopher D. Manning and Jeffrey Heer,\\Stanford University}
\date{Winter 2015}

\begin{document}

\maketitle

\begin{abstract}
``Keyword Extraction'' refers to the task of automatically identifying the most relevant and informative phrases in natural language text. As we are deluged with large amounts of text data in many different forms and content -- emails, blogs, tweets, Facebook posts, academic papers, news articles -- the task of ``making sense'' of all this text by somehow summarizing them into a coherent structure assumes paramount importance. Keyword extraction -- a well-established problem in Natural Language Processing -- can help us here. In this report, we construct and test three different hypotheses (all related to the task of keyword extraction) that take us one step closer to understanding how to meaningfully identify and extract ``descriptive'' keyphrases. The work reported here was done as part of replicating the study by Chuang et al. \cite{Chuang:2012:LCU:2362364.2362367}.
\end{abstract}

\section{Introduction}
\label{sec:intro}

Keyword extraction has been a long-standing problem in Natural Language Processing and Information Retrieval where the goal is to identify (and rank) a set of phrases given a natural language document that best describe the contents of that document in a coherent manner. Usually, a \emph{budget} is imposed on the number of keywords\footnote{In this report, we use the terms ``keywords'' and ``keyphrases'' interchangeably.} to be returned, and the length of the phrases in number of words. Since a document usually talks about several different ideas, topics, and thoughts, potentially interweaving them into a single, coherent body of text, it becomes hard to disentangle it and sieve out only the most important phrases, and still respecting the length and number budgets as described.

When faced with such a task, it is instructive to look into how human beings solve this challenging optimization problem. It has been generally observed that human annotators tend to select certain topics first from the set of topics they feel best describe a given document, and then they pick out phrases that they feel best describe a given topic \cite{LOZA14.1037}. This two-step procedure yields a set of phrases that are both ``locally coherent'' (in that they represent the topical structure of a document), and ``globally adequate'' (in that they are representative of the whole rather than a part of the document).

However, given that human judgment is rather subjective, and the length and number budgets so restrictive, it often becomes the case that given the same piece of text, two annotators come up with two radically different sets of keyphrases -- leading to a very low inter-annotator agreement (20-30\% F-score) \cite{LOZA14.1037}. While this paper suggests ``relaxed matching'' as a way to counteract and alleviate this problem, the inherent subjectivity of keyword extraction has long been a challenge \cite{Kim:2010:STA:1859664.1859668}.

This paper addresses the subjectivity and budget issues of keyword extraction by attacking the problem from an information visualization perspective. The authors -- coming from two different backgrounds (Natural Language Processing and Human-Computer Interaction) -- define ``descriptive keyphrases'' as phrases that best visualize a given document in terms of \emph{tag clouds}.\footnote{See, e.g., \url{http://www.wordle.net/}.} If, among two sets of keyphrases, one has better visualization capability than the other, then the former is regarded as a ``better set of keyphrases''. Note that ``good visualization'' is a subjective term itself, and evaluation of visualizations must somehow take into account that subjectivity.

The key contribution of this paper lies in using online surveys to gather data from many human annotators regarding what they think would constitute valid keyphrases for a document,\footnote{The chosen domain was scientific abstracts, but the general methodology would work in any other domain.} then selecting factors that were most prevalent in human selection of keyphrases, constructing statistical models using those factors to classify phrases as ``key'' vs ``non-key'', and finally evaluating tag clouds (generated by the statistical models) by Amazon Mechanical Turk.\footnote{\url{https://www.mturk.com/mturk/welcome}.} This process of systematic data collection, modeling, and evaluation gets rid of the subjectivity problem by collecting data from \emph{a lot} of annotators instead of a few (as is customary in keyword extraction\footnote{For example, the biggest annotation effort in academic keyword extraction -- the SemEval 2010 Keyphrase Extraction Task -- used 50 annotators \cite{Kim:2010:STA:1859664.1859668}, whereas the current paper uses 69. The more annotators you have, the more confidence you can expect in the obtained keyphrases.}). The authors also counteracted the two budget problems (length and number of keyphrases) by allowing annotators to select as many keyphrases as they wanted (minimum of five), and as long as they wanted. They were further allowed to choose their own vocabulary in building the keyphrases. This is what is known in Natural Language Processing as \emph{abstractive summarization}, where you summarize a document's content using your own words rather than using words/sentences from the document itself (\emph{extractive summarization}).

The authors analyzed human-generated keyphrases to identify several interesting patterns. They showed that bigrams (two-word phrases) were most commonly chosen by humans as keyphrases, followed by unigrams, followed by n-grams with n $\geq$ 2. It was further shown that phrases that were not too rare and not too common, were most frequently chosen as keyphrases, and the majority of keyphrases were technical terms that followed certain part-of-speech patterns. We shall come back to and examine some of these findings in more detail in Section \ref{subsec:exploratory_analysis}.

This report is organized as follows. We describe and explain our research hypothesis in Section \ref{sec:research_hypothesis}, along with refinements and the final version. The project proper is detailed in Section \ref{sec:project}, where we identify and resolve several challenges, including but not limited to lack of data, annotations, and tools. The results are presented in Section \ref{sec:results}, where we also tie them together with our hypothesis, and show that the latter is indeed true under the assumptions we made. We conclude the report in Section \ref{sec:conclusion}.

\section{Research Hypothesis}
\label{sec:research_hypothesis}

Our initial research hypothesis consisted of four parts:

\begin{em}
\begin{boxed}
\begin{enumerate}
\item Multi-word phrases are better descriptors of a scientific abstract than single words (unigrams); for multiple abstracts, this observation needs to be traded off against the better aggregation afforded by unigrams. ``Better''-ness is measured by analyzing frequency histograms on a 144-abstract corpus.
\item Probabilistic measures such as $G^2$ outperform raw term frequency and \emph{tfidf} in terms of precision and recall on scientific abstracts, when logistic regression is used as classifier.
\item Keyphrase quality improves by including part-of-speech and positional features; parse-tree-based features do not lead to additional improvement. Quality is measured by precision and recall.
\begin{itemize}
\item \textbf{Setting:} logistic regression classifier on scientific abstracts, using frequency, commonness, probabilistic, part-of-speech-based, parse-tree-based, and positional features.
\end{itemize}
\item Reducing redundancy (by coalescing semantically identical phrases into a single phrase) helps identify more meaningful phrases, especially from a visualization perspective. ``Meaningful''-ness is measured by average user satisfaction on Amazon Mechanical Turk.
\end{enumerate}
\end{boxed}
\end{em}

These four parts -- we could call them ``four hypotheses'' -- are distilled from the authors' recommendations in Section 8.1 of the paper. We shall first explain the background in more detail, before getting into refinements, and the final version of the hypothesis.

Keyphrase extraction can be tackled as an unsupervised (ranking) problem, or as a supervised (classification) problem. In the unsupervised setting, a set of candidate phrases is first extracted from a document (or a set of documents), and then ranked by a graph-based ranking model (e.g., TextRank \cite{mihalcea-tarau:2004:EMNLP}) or an informativeness-based model \cite{tomokiyo2003language}. Since the ranking model cannot control the number of returned keyphrases (usually a heurtistic cutoff is set at about five or ten top-ranked phrases), it becomes critical that the initial keyphrase selection be as good (and picky) as possible. The selection phase is guided by named entity recognition \cite{Finkel:2005:INI:1219840.1219885,rennie2005using} and noun phrase chunking, and often accompanied by part-of-speech patterns \cite{Hulth:2003:IAK:1119355.1119383}. While noun phrases and named entities can by no means define the complete set of keyphrases for a document (see, e.g., Section \ref{subsec:exploratory_analysis}), they do help in identifying potentially good candidate phrases and filtering out obviously incorrect choices early on in the process.

In \emph{supervised keyword extraction}, on the other hand, candidate phrases are classified as \emph{keyphrase} or \emph{not}. More accurately, a candidate phrase is classified as a \emph{valid keyphrase} for a document vs. \emph{not valid}. Note that this latter setting allows the use of document-based features (e.g., positional, frequency-based, and grammatical) that are not possible in the former setting. Also to be noted is the fact that although most studies in supervised keyphrase extraction treat the problem as binary classification, a few used \emph{sequence labeling} methods such as Conditional Random Fields (CRF) to achieve good performance \cite{zhang2008automatic}. Unsupervised phrase ranking functions (e.g., phrase length, phrase centrality in a network, informativeness) often serve as good features in supervised keyphrase extraction.

Both supervised and unsupervised keyphrase extraction need a final filtering and redundancy reduction step to cope with semantically related phrases that are generated as part of the process. Unsupervised approaches are particularly susceptible to generating keyphrases that are semantically related (often due to poor candidate selection), such as ``Obama'' and ``Barack Obama'' \cite{www200967}.

In this paper, keyphrase extraction is tackled as a supervised problem (i.e., binary classification). Six different types of features were used (frequency, commonness, probabilistic, part-of-speech-based, parse-tree-based, and positional; see Section \ref{subsec:feature_extraction}). The first three types of features need what is known as a \emph{background corpus} in keyphrase extraction literature. The document(s) we are extracting keywords from constitute(s) the \emph{foreground corpus}, whereas the documents being used to extract certain categories of features, are called the \emph{background corpus}. This paper used 144 Ph.D. thesis abstracts from Stanford University as its foreground corpus, and 9,068 Ph.D. theses (presumably abstracts only) as its background corpus. A second variant of commonness features was extracted on a separate background corpus -- Web1T \cite{BrantsFranz:06}.

The usage of Ph.D. thesis abstracts as foreground corpus characterizes and re-inforces a long line of research in \emph{academic keyphrase extraction}. The reason academic keyphrase extraction has received more attention in the literature than other domains, is mostly because of data availability. There are three separate academic keyword extraction datasets (cf. \cite{DBLP:journals/corr/LahiriCC14}, Section 4), whereas there is only one dataset for keyword extraction from meeting transcripts, and one dataset for keyword extraction from news articles (cf. \cite{hasan2010conundrums}, Section 2). The SemEval 2010 Keyphrase Extraction task was specifically devoted to scientific articles \cite{Kim:2010:STA:1859664.1859668}.

The authors of this paper created their own dataset from Stanford thesis abstracts. They looked into both single-document as well as multi-document keyphrase extraction, the latter being tried for the first time in keyword extraction literature (as far as we know). Another novel aspect of this paper is the authors' use of \emph{commonness} as a feature in keyword extraction. Commonness quantifies how ``popular'' or ``rare'' a particular phrase is with respect to the background corpus. Usage of Amazon Mechanical Turk in keyphrase validation is a novelty as well, but that was inspired by the fact that human-computer interfaces (tag cloud visualizations in this case) ultimately need to be evaluated by human annotators.

Keyword extraction systems are often evaluated (and compared) using area under the precision-recall curve, and this paper follows the same approach. A second approach is to use the F-score (harmonic mean of precision and recall) at different cutoff points, and see which system gives better F-scores. The best keyphrase extraction systems in the SemEval 2010 Shared Task achieved F-scores (at cutoff 15) between 20\% and 30\%, by using several complex features, external knowledge bases such as Wikipedia, and sophisticated classifiers such as multi-layered perceptrons and bagged decision trees \cite{berend-farkas:2010:SemEval,Kim:2010:STA:1859664.1859668,Lopez:2010:HAK:1859664.1859719}. This paper on the other hand uses a much smaller and simpler set of features with logistic regression classifier, and shows that the generated keyphrases can still be very good from a visualization standpoint. We shall describe the features in Section \ref{subsec:feature_extraction}.

We refined and qualified our initial hypothesis in light of our revelations from studying the paper in depth, and our experimental findings. We also abandoned the fourth part of the hypothesis because it was more related to visualization and Human-computer Interaction than Natural Language Processing. In the end, our final hypothesis looks as follows:

\begin{em}
\begin{boxed}
While doing single-document supervised keyphrase extraction from scientific abstracts:
\begin{enumerate}
\item Two-word phrases (bigrams) are most frequently chosen by humans, followed by unigrams, trigrams, and higher-order n-grams.
\item Log-odds Ratio outperforms raw term frequency and \emph{tfidf} in terms of area under the precision-recall curve, when logistic regression is used as classifier.
\item Area under the precision-recall curve improves by including part-of-speech and positional features; parse-tree-based features lead to very little additional improvement when part-of-speech features are included.
\begin{itemize}
\item \textbf{Setting:} frequency, commonness, probabilistic, part-of-speech-based, parse-tree-based, and positional features; logistic regression classifier.
\end{itemize}
\end{enumerate}
\end{boxed}
\end{em}

Note that the initial hypothesis included a multi-document keyphrase extraction component. We abandoned it in the final version, because grouping documents into 48 topically similar clusters (with exactly three abstracts in each) proved to be a very challenging optimization problem.\footnote{Please see the following links for discussion on this issue: \url{http://stats.stackexchange.com/questions/8744/clustering-procedure-where-each-cluster-has-an-equal-number-of-points}, \url{http://stackoverflow.com/questions/5452576/k-means-algorithm-variation-with-equal-cluster-size}.} We also abandoned a part of the paper (not the hypothesis) that related extracted keywords to the annotators' \emph{level of familiarity} with the scientific topic, because it was very difficult to control for annotators' level of familiarity on Amazon Mechanical Turk without running into privacy and adversarial issues. Details on our annotation study appear in Section \ref{subsec:data}.

\section{Project}
\label{sec:project}

In this section we shall describe the steps we followed to replicate the study, and gather evidence in favor of our hypothesis. The steps can be broadly classified into three categories -- data collection and annotation, feature extraction, and performance evaluation -- as detailed in the three following subsections.

\subsection{Data Collection and Annotation}
\label{subsec:data}

A big part of any empirical project (all Natural Language Processing tasks included) is to collect data, and this paper is no exception. Recall from Section \ref{sec:research_hypothesis} that we needed two different datasets (\emph{corpora}) -- a foreground corpus, and a background corpus. The authors used Stanford thesis abstracts for both purposes, and Web1T as a second background corpus \cite{BrantsFranz:06}. The problem, however, is that the Stanford dataset was not available to us due to privacy reasons, and Web1T -- although available in our research group -- was too time-consuming to use in an efficient manner (owing to the large number of Web1T queries we needed to run for each experiment).

In the end, we resorted to the SemEval 2010 Keyphrase Extraction Task dataset as our foreground corpus \cite{Kim:2010:STA:1859664.1859668}. More specifically, we used the training set of this corpus as the foreground. The training set has 144 academic papers collected from the ACM Digital Library,\footnote{\url{http://dl.acm.org/dl.cfm}.} from four 1998 ACM classifications: C2.4 (Distributed Systems), H3.3 (Information Search and Retrieval), I2.11 (Distributed Artificial Intelligence -- Multiagent Systems), and J4 (Social and Behavioral Sciences -- Economics). The training set is balanced among the four classes. To make the setup similar to the original paper (which used 144 thesis abstracts as the foreground corpus), we collected the title and abstract of these 144 SemEval papers from ACM Digital Library and copied them into text files.

Our background corpus comes from the \textbf{Inspec} database of 2,000 abstracts -- first described by Anette Hulth, then at Stockholm University \cite{Hulth:2003:IAK:1119355.1119383}. The abstracts are from 1998 to 2002, from journal papers, and from the disciplines \emph{Computers and Control}, and \emph{Information Technology}. The set of abstracts has associated with them two distinct sets of keywords assigned by a professional indexer -- one comprising \emph{controlled terms} (terms that were present in a thesaurus), and another comprising any terms (\emph{uncontrolled terms}) that the indexer deemed appropriate. In our experiments, we only considered the text portion of the abstracts as our \emph{background corpus}, and ignored the keywords because they were not relevant to feature extraction.

The reason we chose Inspec as our background corpus is because it consists of scientific abstracts that are similar in content and length to the SemEval abstracts (the \emph{foreground}). Furthermore, the corpus is large enough to serve as background (i.e., has enough information to gather term statistics and extract features), and yet small enough to be quickly indexed and processed without the overhead of an online query mechanism -- as would be the case with Web1T \cite{BrantsFranz:06}. Furthermore, Inspec has been used as a benchmark in several previous keyword extraction studies \cite{hasan2010conundrums,DBLP:journals/corr/LahiriCC14,mihalcea-tarau:2004:EMNLP}.

Since we did not have access to the original keyword annotations (again, due to privacy reasons), we resorted to Amazon Mechanical Turk to annotate the foreground corpus (144 abstracts) with keyphrases.\footnote{Code and data can be found at \url{http://web.eecs.umich.edu/~lahiri/replication_of_keyword_extraction_part_of_the_paper_by_Chuang_etal_data_and_code.zip}.} The original study collected 69 human-participant responses, so we went ahead and made 69 assignments on Mechanical Turk for each foreground abstract. In Mechanical Turk, every abstract is associated with a \emph{human intelligence task} (HIT) that needs to be completed 69 times (called \emph{assignments} in Mechanical Turk parlance) by 69 different people. In the end, we received 9,936 assignments in total, submitted by 214 unique Turkers. Each assignment consisted of five to 16 keyphrases that could be either abstractive or extractive with no length restrictions imposed on them (in keeping with the spirit of the original study). In total, we received 17,640 unique keyphrases (15,187 after lowercasing).\footnote{Contrast this with 5,611 responses received from 69 human participants in the original study.}

The reason we received so many responses, is because Turkers are creative and they come up with new phrases to describe a document (esp. abstractive phrases). We suspect that Turkers are more creative than the human judges of the original paper -- potentially owing to the fact that the original judges were more educated on average, and therefore, more familiar with academic abstracts. Note that we enforced quality control by making sure that only those Turkers who completed 200 or more HITs with 95\% or higher approval rate could work on our assignments. But even with quality control, Turkers seemed to select arbitrarily long phrases (even sentences and sentence fragments!) to describe abstracts. This indicates a fundamental lack of understanding of the underlying problem. One reason this was the case, is that Turkers were given minimal instructions (in keeping with the spirit of the original study), which probably confused them into over-/under-generating phrases. Verifying this hypothesis is beyond the scope of the present work, but it will be an interesting direction to investigate in future.

Since it was impossible to manually verify all 17,640 unique keyphrases (that was one reason of doing the Mechanical Turk study in the first place), we implemented a simple filter to weed out obviously incorrect or spurious answers.\footnote{We also filtered out ``keyphrases'' that were essentially the whole title of the paper, or the whole abstract -- copied and pasted verbatim.} This filter was constructed manually by looking through responses, and coming up with a list of ``spurious phrases'' (cf. Table \ref{tab:spurious}). After filtering, we are left with 17,621 unique keyphrases (15,173 if lowercased) -- still too large in comparison with the original study. We further restricted our attention to only those phrases that are of length five words or shorter (in keeping with the original), thus ending up with 12,810 keyphrases. Focusing on the phrases that were \emph{extractive} (i.e., the ones that appeared in their corresponding abstracts), we finally obtained 7,816 keyphrases. These 7,816 phrases are what will be used in our supervised keyword extraction experiments (Section \ref{subsec:performance_evaluation}).\footnote{Contrast this with the 2,882 keyphrases in Section 4 of the original study.}

\begin{table}
\begin{center}
%\scriptsize
\begin{tabular}{|l|}
\hline
emptyanswer\\
na\\
nothing\\
optional\\
aa\\
keyword\\
Keyword/Keyphrase 1:  Keyword/Keyphrase 1:\\
Keyword/Keyphrase 1:\\
NA\\
N/A\\
n/a\\
N/a\\
N\textbackslash A\\
Keyword/Keyphrase 1:  vvvvvvvvvvvvvvv\\
Keyword/Keyphrase 1:  vvvvvvvvvvvvvvvvvvvvvvvvvvvvv\\
Keyword query\\
Keywords\\
Keywords in a search form\\
keywords, titles, and full-text.\\
vvvvvvvvKeyword/Keyphrase 1:\\
\hline
\end{tabular}
\caption{Spurious phrases. Note that these are often copied and pasted verbatim from the instructions, and rank among the most frequent responses.}
\label{tab:spurious}
\end{center}
\end{table}

One point of departure from the original study is that the authors mentioned performing \emph{shallow stemming}. However, the extent to which stemming was applied was never discussed in detail. Personal email communication with the first author resulted in the realization that this study was performed eight years back, and many details were lost, thus potentially relegating most of the future experimental decisions to ``educated guesses''. Since the extent of stemming could not be decided in advance, we chose not to stem our keyphrases.

\subsection{Feature Extraction}
\label{subsec:feature_extraction}

We built logistic regression models on the 7,816 human-annotated keyphrases. These 7,816 serve as \emph{positive examples} to the classifier. We obtained 78,160 negative examples by randomly sampling n-grams (1 $\leq$ n $\leq$ 5) from the SemEval corpus without replacement. These negative examples were each given a weight of 0.1 in accordance with the original study.\footnote{The rationale behind choosing ten times as many negative examples as there are positives and then under-weighting the former, was never explained in the paper.} Four types of features -- frequency, commonness, grammatical, and positional -- were extracted for each example phrase (positive and negative). Note that probabilistic features ($G^2$, BM25, weighted log-odds ratio) are subsumed under frequency features, and grammatical features consist of part-of-speech features and parse-tree-based features.

\begin{table}
\begin{center}
%\scriptsize
\begin{tabular}{lc}
\hline
Frequency Feature & Definition\\
\hline
log(tf) & log($t_{Doc}$)\\
&\\
tf.idf & $\frac{t_{Doc}}{t_{Ref}}$ log($\frac{N}{D}$)\\
&\\
$G^2$ & 2($t_{Doc}$ log($\frac{t_{Doc}T_{Ref}}{T_{Doc}T_{Doc}}$) + $t_{\overline{Doc}}$ log($\frac{t_{\overline{Doc}}\ T_{Ref}}{T_{\overline{Doc}}\ T_{Doc}}$))\\
&\\
BM25 & $\frac{3 t_{Doc}}{t_{Doc}\ +\ 2(0.25\ +\ 0.75 \frac{T_{Doc}}{r})}$ log($\frac{N}{D}$)\\
&\\
WordScore & $\frac{t_{Doc}\ -\ t_{Ref}}{T_{\overline{Doc}}\ -\ T_{\overline{Ref}}}$\\
&\\
Weighted log-odds ratio & $\frac{\log(\frac{t'_{Doc}}{t'_{\overline{Doc}}})\ -\ \log(\frac{T'_{Doc}}{T'_{\overline{Doc}}})}{\sqrt{\frac{1}{t'_{Doc}}\ +\ \frac{1}{t'_{\overline{Doc}}}}}$\\
\hline
\end{tabular}
\caption{Frequency features. Note that we dropped WordScore because the definition of $T_{\overline{Ref}}$ was not clear from the paper. The other definitions are explained as follows. Given a document from a reference corpus with $N$ documents (in our case, Inspec with 2,000 documents), the score for a term is given by these formulas. $t_{Doc}$ and $t_{Ref}$ denote term frequency in the document and reference corpus; $T_{Doc}$ and $T_{Ref}$ are the number of \emph{words} in the document and reference corpus; $D$ is the number of documents in which the term appears; $r$ is the \emph{average word count} per document; $t'$ and $T'$ indicate measures for which we increment term frequencies in each document by 0.01; terms present in the corpus but not in the document are defined as $t_{\overline{Doc}} = t_{Ref} - t_{Doc}$ and $T_{\overline{Doc}} = T_{Ref} - T_{Doc}$. Among the family of tf.idf measures, a \emph{reference-relative form} was chosen as shown. For BM25, the parametrization of $k_1 = 2$ and $b = 0.75$ has been suggested in previous literature. A \emph{term} is any analyzed phrase (n-gram). When frequency statistics are applied to n-grams with n = 1, the terms are all the individual words in the corpus. When n = 2, scoring is applied to all unigrams and bigrams in the corpus, and so on.}
\label{tab:freq}
\end{center}
\end{table}

Frequency features have been summarized in Table 3 of the original paper, which we reproduce here for conceptual clarity (Table \ref{tab:freq}). Note that although frequency features are often used in keyphrase extraction, their original intent was to help assess \emph{document relevance} in information retrieval. Note also that probabilistic features -- $G^2$, BM25, weighted log-odds ratio -- are not used in keyphrase extraction that often. We will later show that one probabilistic feature -- weighted log-odds ratio -- outperforms other frequency-based features (Section \ref{subsec:model_performance}).

Commonness features (briefly mentioned in Section \ref{sec:research_hypothesis}) encode the \emph{relative rarity} of a phrase against a background corpus. \emph{Commonness} of a term (n-gram) is defined as $\frac{log(tf_{bg})}{log(tf_{max})}$, where $tf_{bg}$ is the frequency of the term in the background corpus, and $tf_{max}$ is the frequency of the most frequent n-gram (with the same number of words as the term) in the background corpus. The authors used two background corpora to extract commonness information -- Stanford thesis abstracts and Web1T, and experimented with up to 20 \emph{bins} of commonness (which we shall replicate in Section \ref{subsec:model_performance}). We used \textbf{Inspec} as our sole background corpus, so the commonness information comes from Inspec.

We extracted ten binary grammatical features (yes/no) -- six of them based on parse trees, and rest based on parts-of-speech.\footnote{Note that it was not clear from the paper whether these features were binary or categorical, but based on further pondering, it seemed to us that binary will be the best choice.} The Stanford Parser was used to extract grammatical features \cite{klein-manning:2003:ACL}. We implemented several Java functions within Stanford Parser code to make it work according to our specifications. The grammatical features are as follows (first six are parse-tree-based, and the rest are part-of-speech-based):

\begin{itemize}
\item \textbf{Is full noun phrase?} If a phrase matches a full noun phrase, we mark this feature as a one; otherwise, it is zero.
\item \textbf{Is full verb phrase?} If a phrase matches a full verb phrase, we mark this feature as a one; otherwise, it is zero.
\item \textbf{Is partial noun phrase?} If a phrase is part of a full noun phrase, we mark this feature as a one; otherwise, it is zero.
\item \textbf{Is partial verb phrase?} If a phrase is part of a full verb phrase, we mark this feature as a one; otherwise, it is zero.
\item \textbf{Is optional leading word?} If the first word of a phrase is an \emph{optional leading word} (leading word in a noun phrase whose part-of-speech is a cardinal number (CD), a determiner (DT), or a pre-determiner (PDT)), we mark this feature as a one; otherwise, it is zero.
\item \textbf{Is head noun?} If the last word of a phrase is a \emph{head noun} (last word of a noun phrase), we mark this feature as a one; otherwise, it is zero.
\item \textbf{Is technical term?} If a phrase matches one of the part-of-speech patterns for a technical term, we mark this feature as a one; otherwise, it is zero.
\item \textbf{Is compound technical term?} If a phrase matches one of the part-of-speech patterns for a compound technical term, we mark this feature as a one; otherwise, it is zero.
\item \textbf{Is partial technical term?} If a phrase is part of a technical term, we mark this feature as a one; otherwise, it is zero.
\item \textbf{Is partial compound technical term?} If a phrase is part of a compound technical term, we mark this feature as a one; otherwise, it is zero.
\end{itemize}

Technical terms and compound technical terms are identified by the following regular expression productions on parts-of-speech:

\begin{equation}T = (A|N)^+\ (N|C)\ |\ N\end{equation}
\begin{equation}X = (A|N)^*\ N\ of\ (T|C)\ |\ T\end{equation}

where $T$ is a technical term, $X$ is a compound technical term, $A$ is an adjective (corresponding to part-of-speech tags JJ, JJR, JJS), $N$ is a noun (corresponding to part-of-speech tags NN, NNS, NNP, NNPS), and $C$ is a cardinal number (corresponding to part-of-speech tag CD).\footnote{Note that all technical terms are compound technical terms, but not the other way around. Note also that $of$ is a terminal in production (2).} All part-of-speech tags are in the standard Penn Treebank format (cf. Table 2 of \cite{Marcus:1993:BLA:972470.972475}).

Finally, we extracted three positional features -- \emph{absolute first occurrence} (real number between 0 and 1), \emph{relative first occurrence} (real number between 0 and 1), and \emph{presence in first sentence} (binary yes/no) -- defined as follows:

\begin{itemize}
\item \textbf{Absolute first occurrence:} Earliest position (i.e., word index) of the document where a phrase first appeared, normalized by the total number of words in the document.
\item \textbf{Relative first occurrence:} $(1-a)^k$, where $a$ is the absolute first occurrence of a phrase, and $k$ is the number of times the phrase appeared in the document. Relative first occurrence penalizes frequently occurring phrases' first occurrences.
\item \textbf{Presence in first sentence:} If a phrase is present in the first sentence, we mark this feature as a one; otherwise, it is zero. We used NLTK for sentence segmentation \cite{Loper:2002:NNL:1118108.1118117}.
\end{itemize}

\subsection{Performance Evaluation}
\label{subsec:performance_evaluation}

Recall from Section \ref{sec:research_hypothesis} that the authors performed \emph{supervised keyphrase extraction} by feeding positive and negative example phrases (along with their features) to a classifier. Upon closer inspection, it seemed to us that what was really done, was \emph{supervised ranking} (i.e., learning to rank) using a regression model rather than supervised classification. This observation could not be confirmed (or otherwise) with the first author, because most details of the original study were lost, leaving us to work with best guesses.

It was clear, however, that logistic regression was used as the learning model, and it also became clear, upon further inspection, that R was used as the library of choice for working with logistic regression models. R has some nice parametrizations to weight each sample differently, and that facility was used to under-weight negative examples (cf. \emph{glm()} in \cite{r_citation}). Furthermore, precision-recall curves seemed also to have been generated using R. In particular, we used the package ROCR to generate precision-recall curves from raw predictions \cite{rocr_citation}.

The choice of using precision-recall curves (i.e., area under the curves) to measure and compare performance of different models may seem arbitrary at first blush, but do note that precision-recall curves show a clear trade-off between retrieving good phrases at top ranks vs. retrieving as many good phrases as possible, no matter what the rank is. This trade-off is not so obvious with other measures of performance such as F-score at rank $k$, or \emph{mean average precision}. Also to be noted is the fact that precision-recall curves have a firm grounding in Information Retrieval community in general, and keyword extraction literature in particular \cite{hasan2010conundrums}.

We experimented with five different logistic regression models -- the first four corresponding (almost) exactly to the four sub-figures of Figure 4 in the original paper, and the last one to assess the impact of commonness bins. The models are constructed on the following features:

\begin{itemize}
\item Frequency and probabilistic features.
\item Adding commonness to raw term frequency, and comparing the joint model with raw term frequency and $G^2$.
\item Adding grammatical features to frequency and commonness features.
\item Adding positional features (to the rest of the features) to create the final models.
\end{itemize}

\section{Results}
\label{sec:results}

We describe the results in two subsections -- the first one is an exploratory analysis on Amazon Mechanical Turk responses, and the second one illustrates the performance of logistic regression models for supervised keyphrase extraction under different feature combinations, using precision-recall curves. We shall also show evidence in these two subsections that goes in favor of our research hypothesis (cf. the final hypothesis in Section \ref{sec:research_hypothesis}), and show that the three parts of our hypothesis are indeed true, on our dataset, annotations, features, and models.

\subsection{Exploratory Analysis}
\label{subsec:exploratory_analysis}

\begin{figure*}
    \centering
    \begin{subfigure}{\textwidth}
        \centering
        \includegraphics[width=\textwidth]{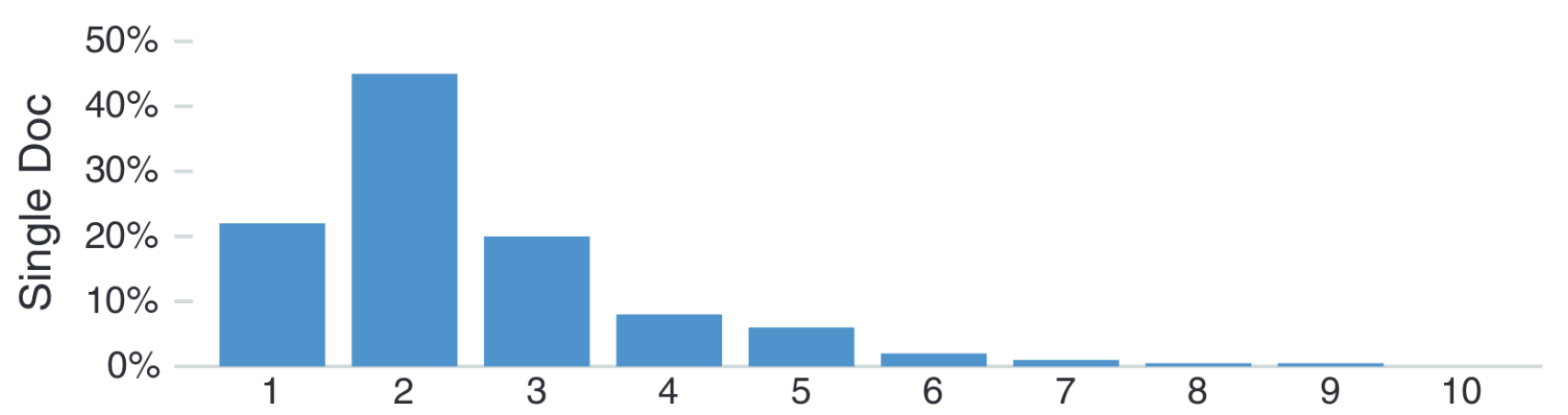}
        \caption{Original}
        \label{fig:phrase_length_original}
    \end{subfigure}

    \begin{subfigure}{\textwidth}
        \centering
        \includegraphics[width=\textwidth]{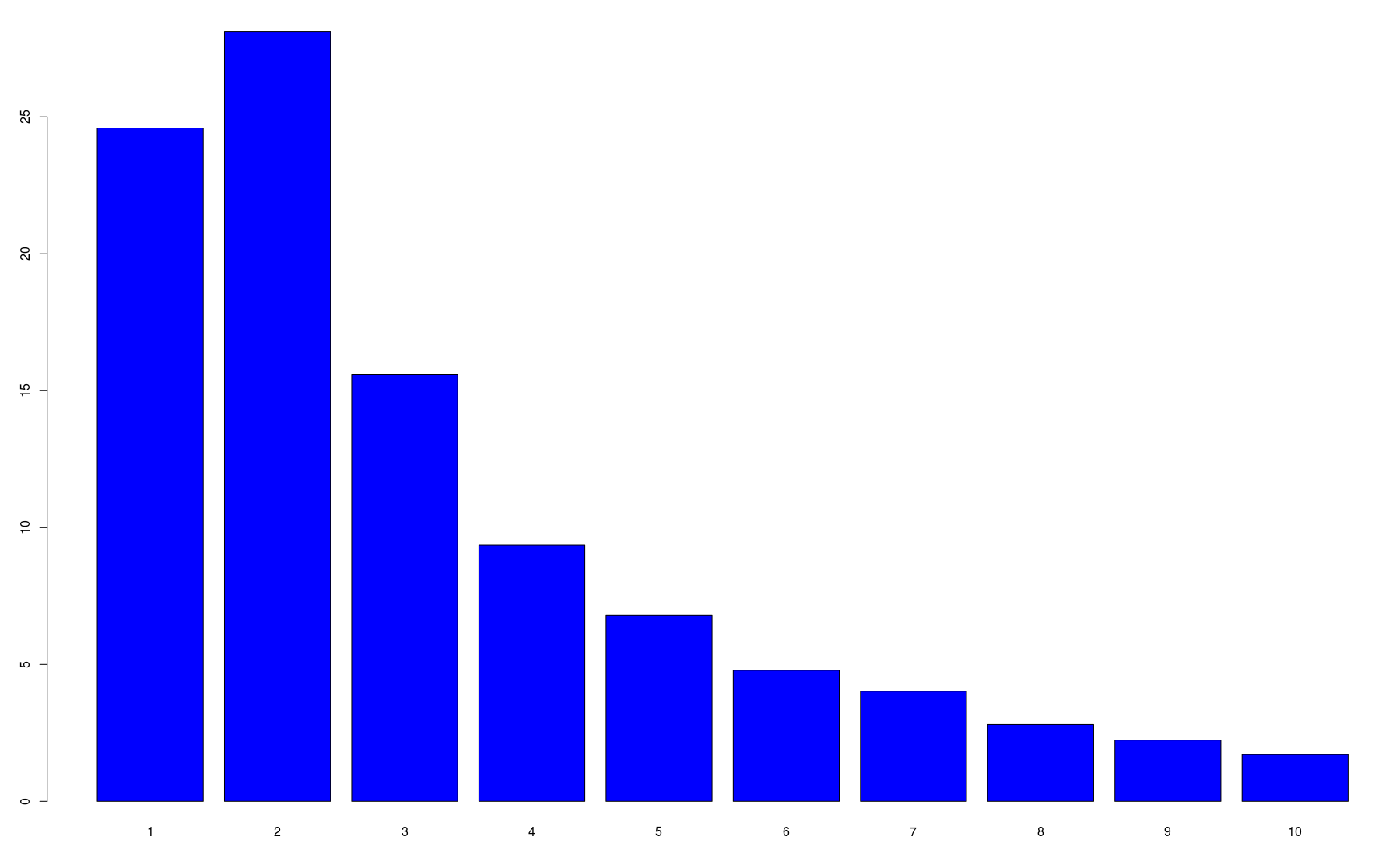}
        \caption{After replication}
        \label{fig:phrase_length_replicated}
    \end{subfigure}
    \caption{Distribution of phrase length. X-axis is the phrase length (in number of words), and Y-axis is the percentage of (human-generated) phrases with that length.}
    \label{fig:phrase_length}
\end{figure*}

The paper spends considerable amount of space to detail the results of their exploratory analysis, for good reasons. \emph{A priori}, it is hard to define or validate keyphrases on scientific abstracts (or on any other domain), let alone ``descriptive'' keyphrases. The authors took the exploratory analysis as an opportunity to look into people's minds to see what they usually think are the most important components of a document. Recall from Section \ref{subsec:data} that the annotation instructions were minimal (``summarize the content using five or more keyphrases, using any vocabulary''). This was done to ensure that people could freely choose what they thought would best describe a given document, without being biased by experimenters' (often arbitrary) decisions.

While the original intent of minimal instructions was clear from the paper, note that we used Amazon Mechanical Turk, which somehow calls for more control on our part. However, we shall see in the remainder of this report that even without sufficiently detailed instructions, we obtained results that matched the hypothesis of the original paper.

As a starter, we looked into the frequency distribution of n-grams in human-generated keyphrases. The paper says: ``bigrams are the most common response, accounting for 43\% of all free-form keyphrase responses, followed by unigrams (25\%) and trigrams (19\%).'' After replication, the numbers we obtained are: 24.5\% bigrams, 21.4\% unigrams, and 13.6\% trigrams. \textbf{This validates the first part of our final hypothesis (Section \ref{sec:research_hypothesis}).} The full distributions are shown in Figure \ref{fig:phrase_length}. There are several points to note here. Figure \ref{fig:phrase_length_original} shows only \emph{one} histogram from the three reported in Figure 2 of the original study, because we did not experiment with \emph{multiple-document keyphrase extraction} (reasons explained in Section \ref{sec:research_hypothesis}). Also to be noted is the fact that we have a much heavier and longer tail in the histogram. For illustration purposes, we capped the tail at 10 words per phrase. In fact, while the original study reports $<$5\% of longer-than-five-words responses, we have 26.45\% of responses that are longer than five words. The reason, as explained in Section \ref{subsec:data}, is that Turkers selected very long phrases (even sentences and sentence fragments) because of our minimal instructions. Another point is that we have a substantially lower peak at bigrams, and a somewhat lower peak at trigrams, the mass being shifted to longer keyphrases (there are many keyphrases longer than 10 words, but we capped them due to visual clarity).

The authors found that 65\% of keyphrase responses were \emph{extractive}, i.e., they were present in the original document. For us, the number was slightly higher: 66.55\%. They also found that ``22\% of keyphrases occur in the first sentence, even though first sentences contain only 9\% of all terms.'' We found that 18.6\% of keyphrases occurred in the first sentence, whereas first sentences contained 25.33\% of all unique terms. To be noted is the fact that our foreground corpus is different than the authors' (SemEval vs. Stanford thesis abstracts), and the former potentially contains smaller abstracts (conference paper abstracts) than theses, thus concentrating many important terms in first sentences.\footnote{Our SemEval corpus contains 6.76 sentences on average (median = 6, standard deviation = 2.42).} Further, the original study reported that nearly ``two-thirds of keyphrases found in the document'' (i.e., \emph{extractive} keyphrases) ``are part of a noun phrase'', ``7\% are part of a verb phrase'', and ``over 80\% \ldots are part of a technical term''.\footnote{Please see Section \ref{subsec:feature_extraction} for the definition of technical terms and compound technical terms.} Note that these numbers add up to more than 100\%, which is fine because many technical terms are actually noun phrases, so they are double-counted. After replication, we found that 38.73\% unique keyphrases are part of a noun phrase, 43.18\% are part of a verb phrase, and 25.62\% are part of a technical term. That we have many more verb phrases, and many fewer noun phrases and technical terms, can be tentatively explained as follows:

\begin{itemize}
\item We provided minimal instructions, thereby confusing Turkers.
\item Turkers on average are less educated than the annotators of the original study, and potentially much less familiar with academic jargon.
\item Turkers over-emphasized the \emph{actions} (verb phrases) rather than \emph{content} (noun phrases and technical terms), perhaps owing to their lack of familiarity with the domain.
\item Our abstracts were shorter than thesis abstracts, and less laden with technical terms.
\end{itemize}

\begin{figure*}
    \centering
    \begin{subfigure}{\textwidth}
        \centering
        \includegraphics[width=\textwidth]{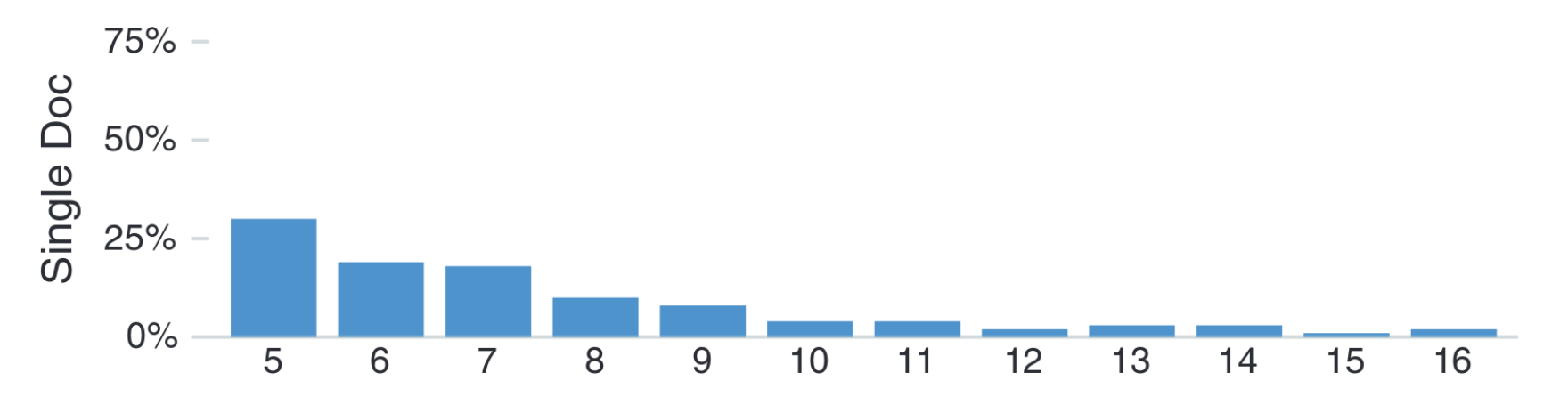}
        \caption{Original}
        \label{fig:number_of_keyphrases_original}
    \end{subfigure}

    \begin{subfigure}{\textwidth}
        \centering
        \includegraphics[width=\textwidth]{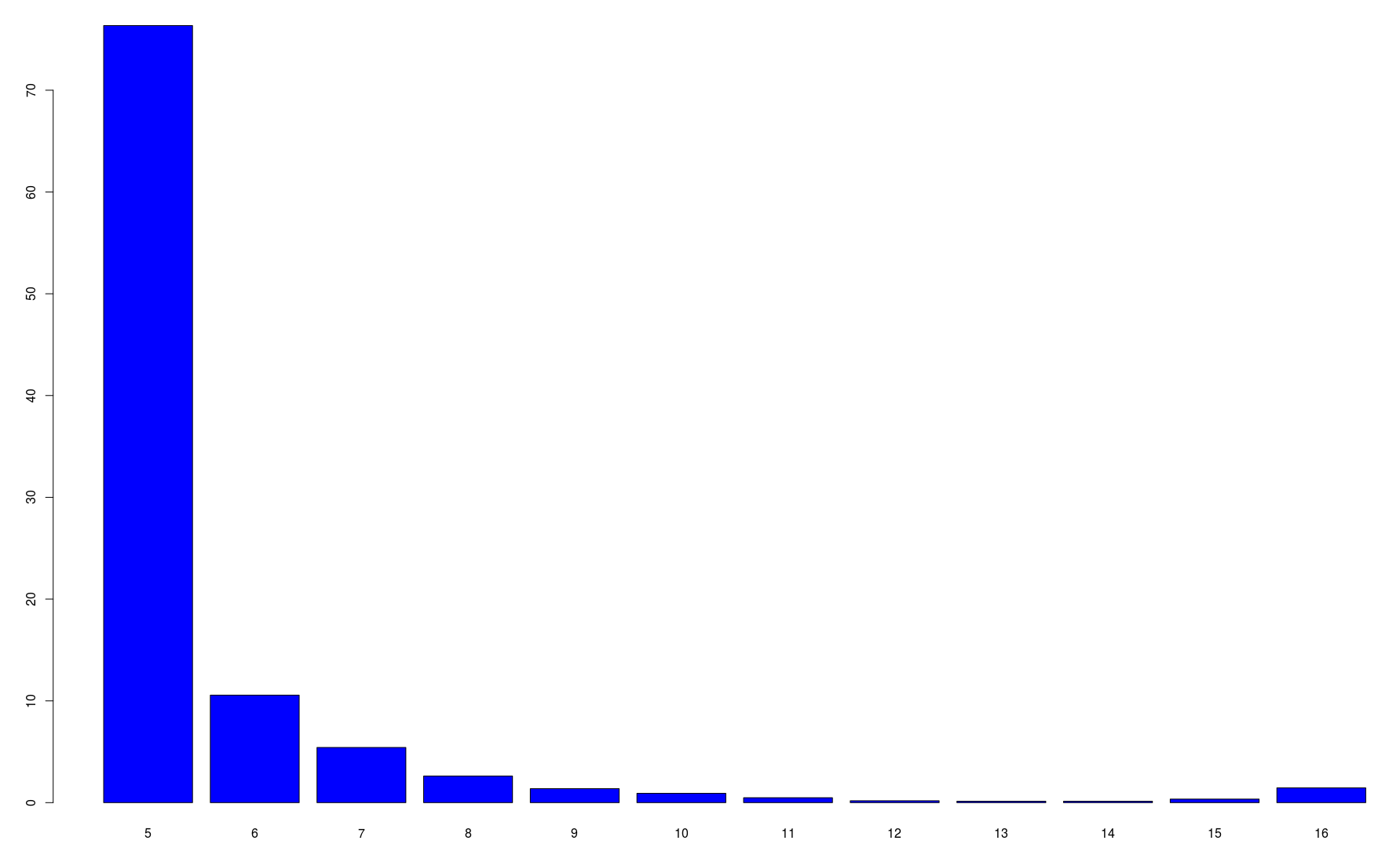}
        \caption{After replication}
        \label{fig:number_of_keyphrases_replicated}
    \end{subfigure}
    \caption{Distribution of the number of human-assigned keyphrases. X-axis is the number of keyphrases, and Y-axis is the percentage of documents/assignments with that number. See text for details.}
    \label{fig:number_of_keyphrases}
\end{figure*}

An interesting aspect of the original study was to see how many keyphrases people use to describe a given scientific abstract. It was observed that for the \emph{single-document case}, most people used five phrases, and the number gradually decreased thenceforth (cf. Figure 1 in the paper), tapering off at 16 phrases. This is why we chose 5-16 as our number of keyphrases in the Mechanical Turk study. In our case, the percentages look as in Figure \ref{fig:number_of_keyphrases_replicated}. Figure \ref{fig:number_of_keyphrases_original} is a duplicate of the first part of Figure 1 of the original. Note that the five-phrase peak is much higher in our case than in the original, and other peaks are much lower. This points to the fact that in most cases, Turkers preferred not to spend any time at all on providing extra information (i.e., optional information). The general shape of the histogram also closely follows the original, thereby validating the authors' comments: ``the peak at five \ldots suggests that subjects might respond with fewer without this requirement. However, it is unclear whether this reflects a lack of appropriate choices or a desire to minimize effort.'' In our case, the much sharper peak at five phrases, and much lower peaks at others suggest that the latter might be the real reason.

Another difference with the original is that since we have a varying number of keyphrases from each assignment (recall from Section \ref{subsec:data} that an \emph{assignment} is a Mechanical Turk term for a single piece of human intelligence task (HIT) to be completed by a single person in a single session), we plotted the histogram in Figure \ref{fig:number_of_keyphrases_replicated} \emph{on the assignments} rather than on the abstracts (which presumably is the case with Figure \ref{fig:number_of_keyphrases_original}), thus potentially inflating the number of five-phrase responses, and deflating others.

\begin{figure*}
    \centering
    \begin{subfigure}{\textwidth}
        \centering
        \includegraphics[width=\textwidth]{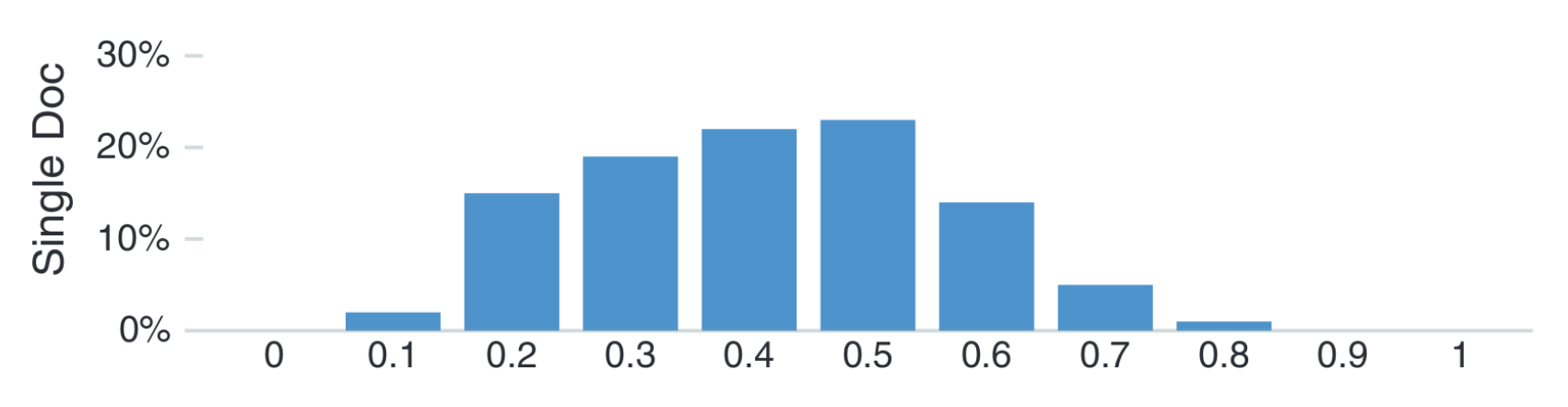}
        \caption{Original (on Web1T)}
        \label{fig:comm}
    \end{subfigure}

    \begin{subfigure}{0.8\textwidth}
        \centering
        \includegraphics[width=\textwidth]{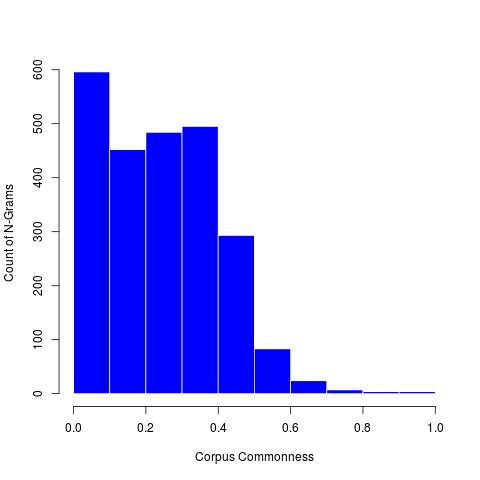}
        \caption{After replication (on Inspec)}
        \label{fig:comm_replicated}
    \end{subfigure}
    \caption{Distribution of commonness. X-axis is commonness, and Y-axis is the percentage/number of unique (human-selected) n-grams with that commonness.}
    \label{fig:commonness}
\end{figure*}

\begin{figure*}
    \centering
    \begin{subfigure}{0.4\textwidth}
        \centering
        \includegraphics[width=\textwidth]{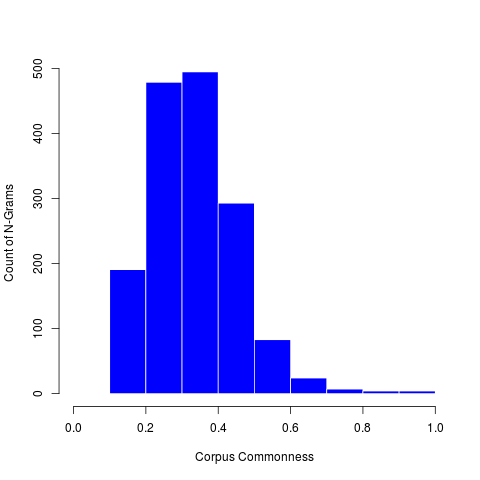}
        \caption{All terms with frequency $<$ 5 are removed.}
        \label{fig:corpus_commonness_histogram_minimum_frequency_cutoff_5}
    \end{subfigure}
    \hfill
    \begin{subfigure}{0.4\textwidth}
        \centering
        \includegraphics[width=\textwidth]{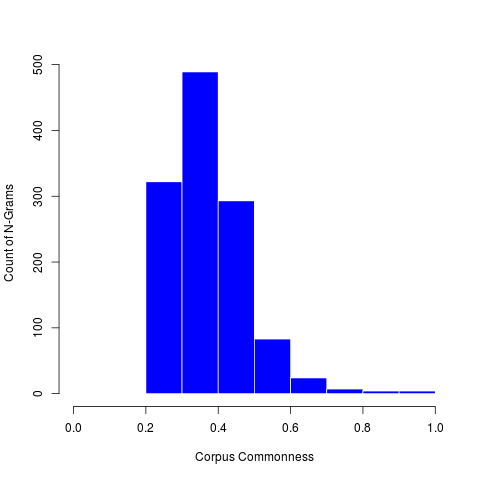}
        \caption{All terms with frequency $<$ 10 are removed.}
        \label{fig:corpus_commonness_histogram_minimum_frequency_cutoff_10}
    \end{subfigure}

    \begin{subfigure}{0.4\textwidth}
        \centering
        \includegraphics[width=\textwidth]{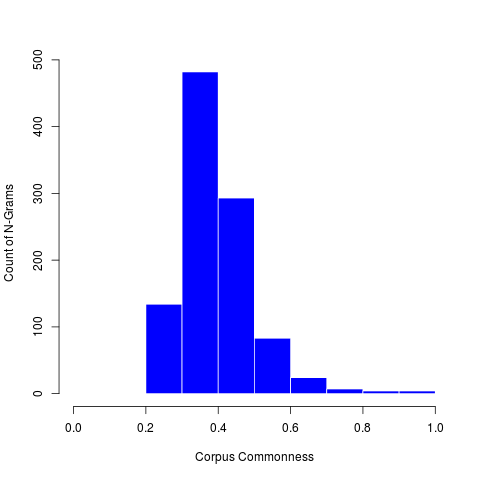}
        \caption{All terms with frequency $<$ 15 are removed.}
        \label{fig:corpus_commonness_histogram_minimum_frequency_cutoff_15}
    \end{subfigure}
    \hfill
    \begin{subfigure}{0.4\textwidth}
        \centering
        \includegraphics[width=\textwidth]{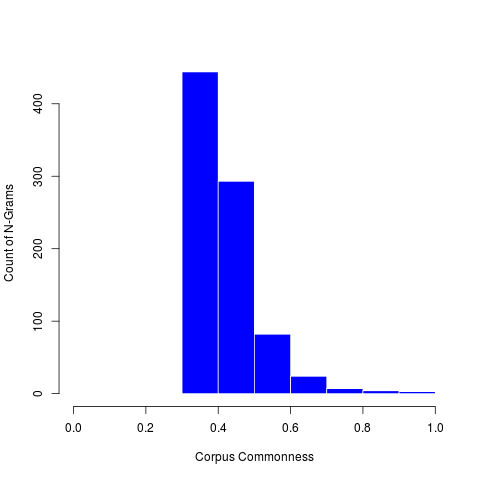}
        \caption{All terms with frequency $<$ 20 are removed.}
        \label{fig:corpus_commonness_histogram_minimum_frequency_cutoff_20}
    \end{subfigure}
    \caption{Distribution of commonness on \emph{Inspec} after the rarest terms are removed. X-axis is commonness, and Y-axis is the number of unique (human-selected) n-grams with that commonness. Note that as we remove more and more rare terms, the figures look more and more similar in shape to Figure \ref{fig:comm} -- but up to a point. Figure \ref{fig:corpus_commonness_histogram_minimum_frequency_cutoff_20}, for example, has removed so many of the rare terms that the general \emph{mid-commonness} shape disappears again.}
    \label{fig:commonness_cutoff}
\end{figure*}

Exploration of \emph{commonness} was an important aspect of the paper.\footnote{Please see Section \ref{subsec:feature_extraction} for the definition of commonness.} The authors found that annotators tend to prefer phrases that are not too rare, and not too common. They called those phrases \emph{mid-commonness}. Validation of this argument came from Figure 3 of the paper, the first part of which is repeated in Figure \ref{fig:comm}. Our replicated version appears in Figure \ref{fig:comm_replicated}. There are several things to note here. First, we computed commonness on our background corpus (\emph{Inspec}) rather than Web1T as was done in the paper (cf. Section \ref{subsec:data}). So, our commonness peaks are much higher than Web1T peaks (because on the Web, which is a huge corpus, all terms are sufficiently rare; whereas on Inspec, which is way smaller, most terms are not very rare). Second, we got a peak -- the largest peak in fact -- at the \emph{lowest} (leftmost) commonness bin. The reason for this is that Web1T excludes extremely rare terms -- terms that appeared less than 40 times \cite{Evert:2010:GWM:1868765.1868770} -- which is why there is no peak at low commonness values. While this does make sense for an enormous background corpus like Web1T, for our much smaller Inspec corpus, deleting such terms does not seem to be worthwhile. Now, given that, it seems to be the case that the authors did miss a point: annotators do not prefer the \emph{mid-commonness terms} as they suggested; they instead prefer the \emph{rarest terms}. Third, we were able to re-construct the pattern of the original histogram (Figure \ref{fig:comm}) by deleting the rarest terms from Inspec corpus (cf. Figure \ref{fig:commonness_cutoff}). This shows that at least for \emph{corpus commonness}, rarest terms are preferred to mid-frequency terms. Of course, if we remove the rarest terms, then the next highest preferred set of terms comes from mid-commonness range.

\subsection{Model Performance}
\label{subsec:model_performance}

\begin{figure}
    \centering
    \begin{subfigure}{0.4\textwidth}
        \centering
        \includegraphics[width=\textwidth]{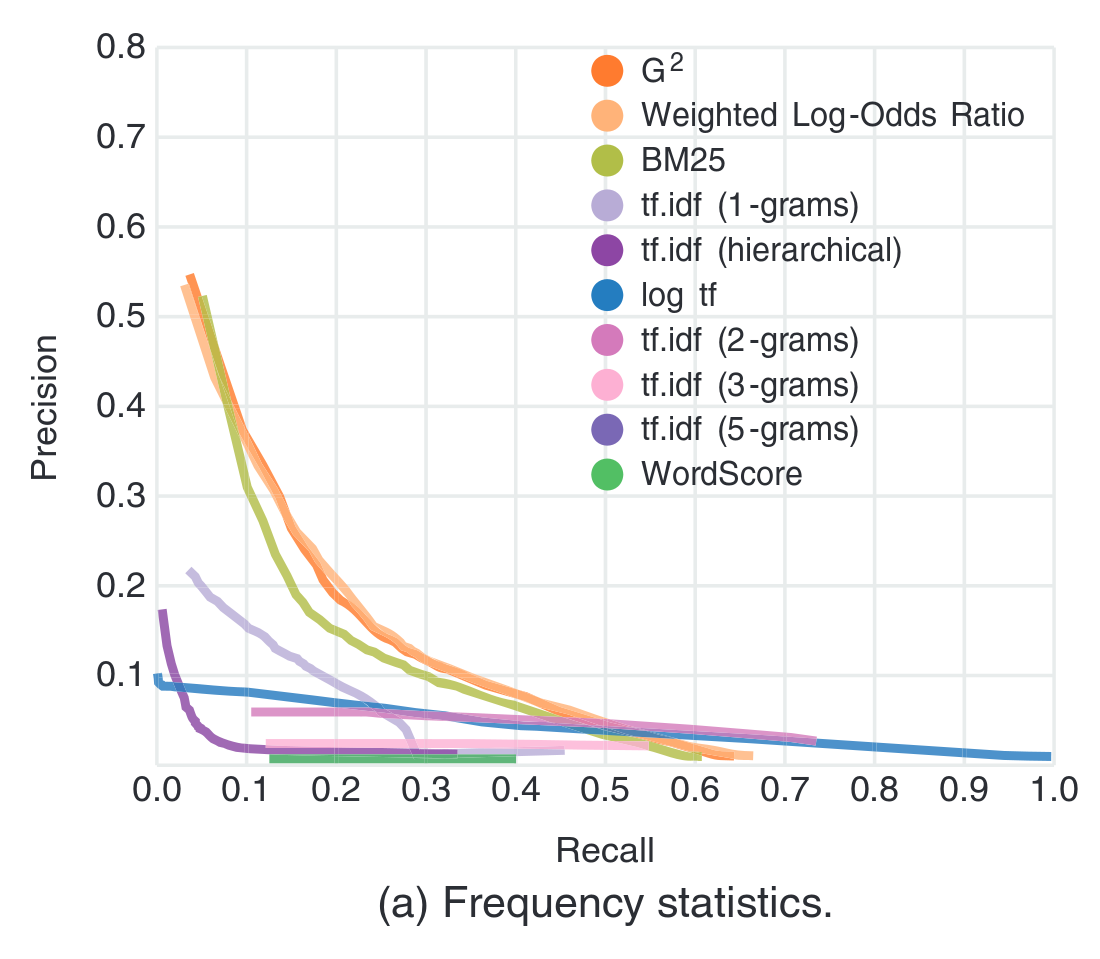}
        \caption{Original}
        \label{fig:1}
    \end{subfigure}
    \hfill
    \begin{subfigure}{0.4\textwidth}
        \centering
        \includegraphics[width=\textwidth]{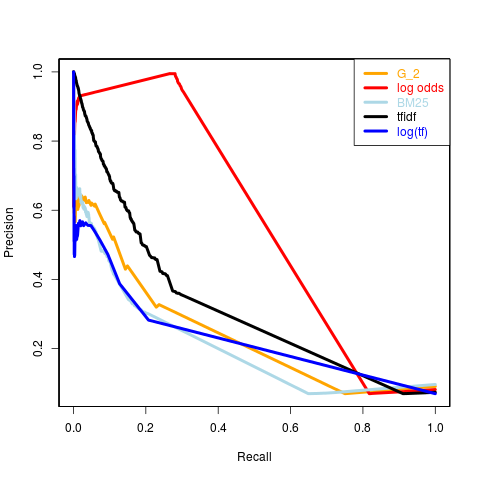}
        \caption{After replication}
        \label{fig:frequency_statistics}
    \end{subfigure}
    \caption{Performance of frequency and probabilistic features (best viewed in color).}
    \label{fig:frequency_features}
\end{figure}

Recall from Section \ref{subsec:performance_evaluation} that we constructed logistic regression models on the following features:

\begin{itemize}
\item Frequency and probabilistic features.
\item Adding commonness to raw term frequency, and comparing the joint model with raw term frequency and $G^2$.
\item Adding grammatical features to frequency and commonness features.
\item Adding positional features (to the rest of the features) to create the final models.
\end{itemize}

Performance was evaluated using precision-recall curves (i.e., area under the curves). In the next set of figures, we shall illustrate the results one by one, along with the original. In the process, we shall also show evidence in favor of the second and third part of our final hypothesis (cf. Section \ref{sec:research_hypothesis}).

Performance of frequency and probabilistic features is given in Figure \ref{fig:frequency_features}. Note that we dropped WordScore (cf. caption of Table \ref{tab:freq}), and did not experiment with the \emph{hierarchical} version of \emph{tfidf}, so we have only five plots in Figure \ref{fig:frequency_statistics} instead of ten in Figure \ref{fig:1}. In Figure \ref{fig:frequency_statistics}, log-odds ratio outperforms log(tf) and \emph{tfidf} in terms of area under the curve. \textbf{This validates the second part of our final hypothesis (Section \ref{sec:research_hypothesis}).} The difference with Figure \ref{fig:1} is that in Figure \ref{fig:1}, probabilistic features ($G^2$, BM25, log-odds ratio) outperformed all other frequency features. This did not happen in our case. We observed that while log-odds ratio gives a very good performance (see the red curve in Figure \ref{fig:frequency_statistics}), other probabilistic features such as $G^2$ and BM25 are much worse -- sometimes even worse than log(tf).

\begin{figure}
    \centering
    \begin{subfigure}{0.4\textwidth}
        \centering
        \includegraphics[width=\textwidth]{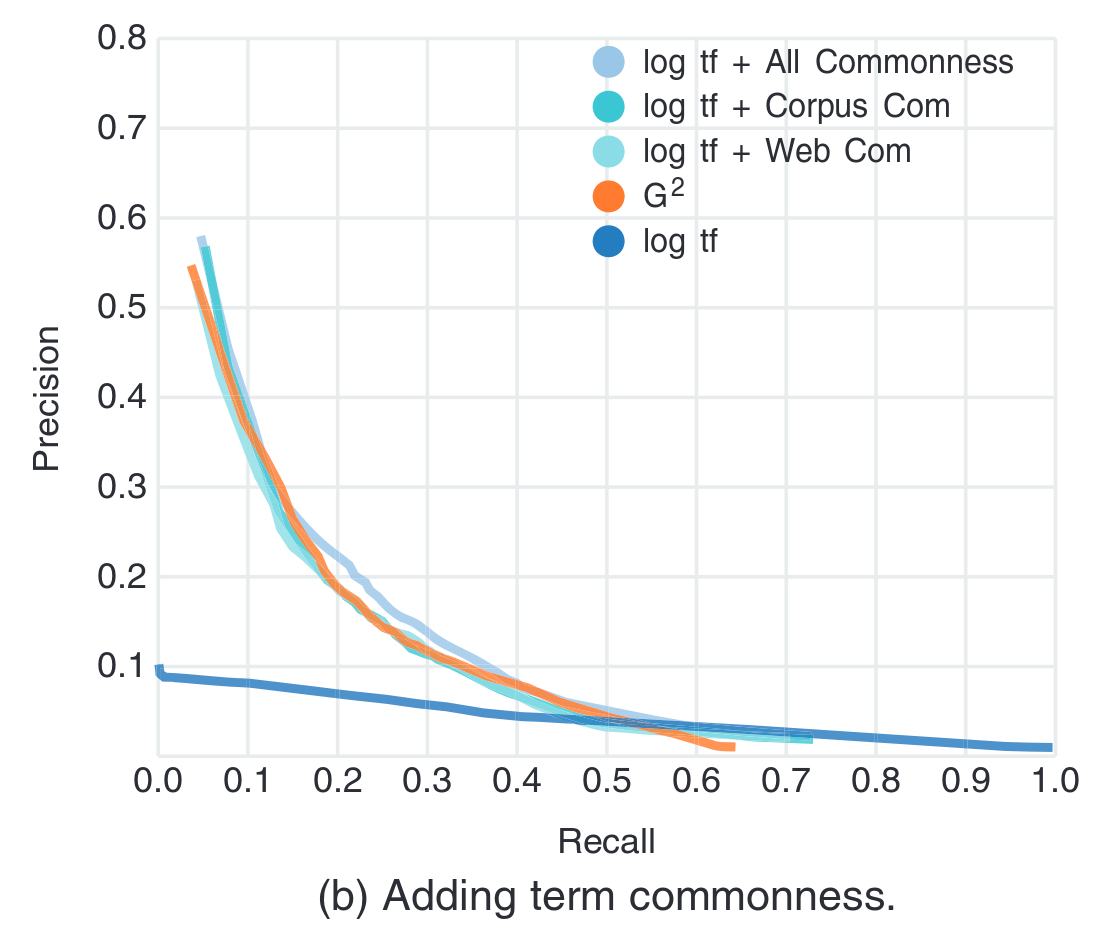}
        \caption{Original}
        \label{fig:2}
    \end{subfigure}
    \hfill
    \begin{subfigure}{0.4\textwidth}
        \centering
        \includegraphics[width=\textwidth]{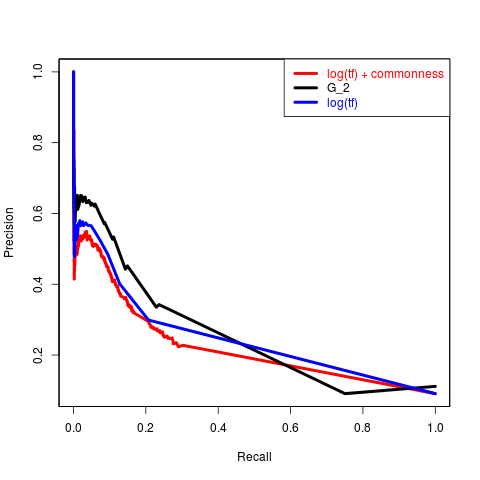}
        \caption{After replication}
        \label{fig:adding_term_commonness}
    \end{subfigure}
    \caption{Adding term commonness to term frequency (best viewed in color).}
    \label{fig:adding_commonness}
\end{figure}

Adding term commonness (\emph{corpus commonness} based on the Inspec background corpus) did not help much in our case (as shown in Figure \ref{fig:adding_term_commonness}), unlike the original version (Figure \ref{fig:2}). Since the original paper experimented with two forms of commonness (web and corpus), they have five plots in total, whereas we have only three.

\begin{figure}
    \centering
    \begin{subfigure}{0.4\textwidth}
        \centering
        \includegraphics[width=\textwidth]{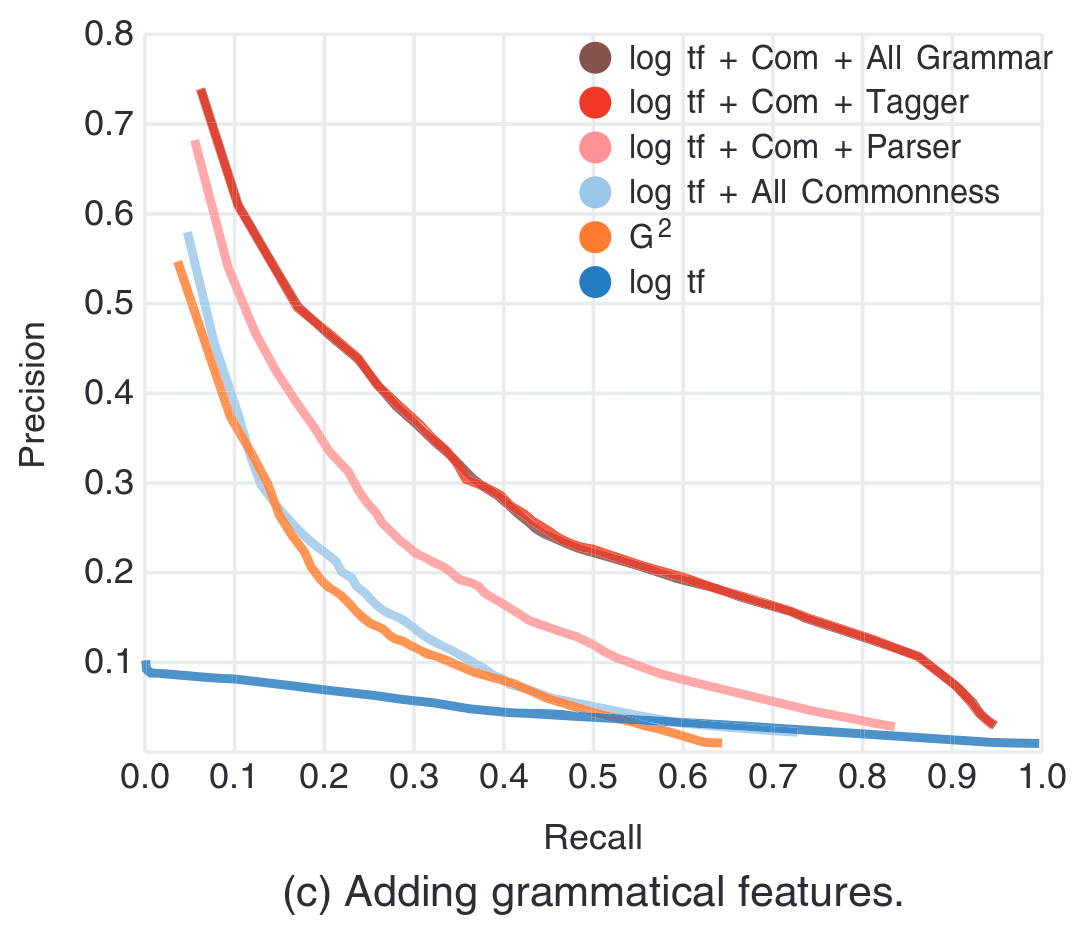}
        \caption{Original}
        \label{fig:3}
    \end{subfigure}
    \hfill
    \begin{subfigure}{0.4\textwidth}
        \centering
        \includegraphics[width=\textwidth]{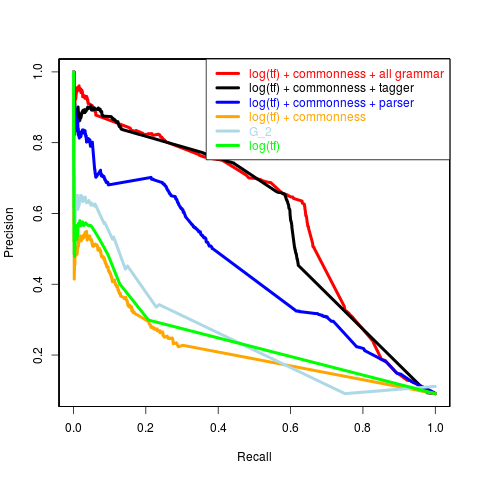}
        \caption{After replication}
        \label{fig:adding_grammatical_features}
    \end{subfigure}
    \caption{Adding grammatical features on top of term commonness (best viewed in color).}
    \label{fig:adding_grammatical}
\end{figure}

Adding grammatical features on top of commonness features, however, gives a big boost in performance. Note from Figure \ref{fig:adding_grammatical_features} that adding grammatical features on top of log(tf) and commonness outperforms log(tf) + commonness combination, log(tf) itself, as well as the probabilistic feature $G^2$. Similar patterns are observed in Figure \ref{fig:3} from the original. Furthermore, the log(tf) + commonness + all grammar combination (red curve in Figure \ref{fig:adding_grammatical_features}) -- which includes parse-tree-based features -- gives very little additional improvement over the black curve (log(tf) + commonness + part-of-speech features). This is again similar to the original. \textbf{This validates the third and final part of our final hypothesis (Section \ref{sec:research_hypothesis}).}

\begin{figure}
    \centering
    \begin{subfigure}{0.4\textwidth}
        \centering
        \includegraphics[width=\textwidth]{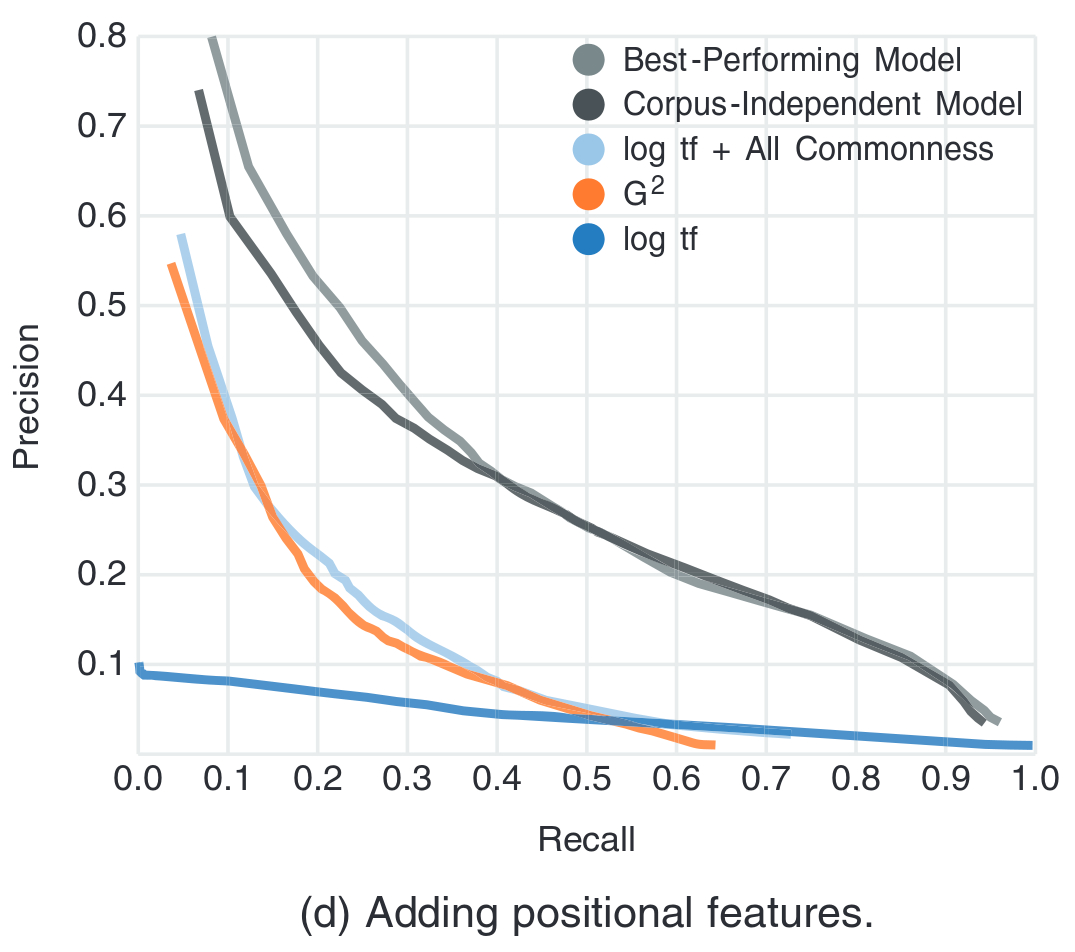}
        \caption{Original}
        \label{fig:4}
    \end{subfigure}
    \hfill
    \begin{subfigure}{0.4\textwidth}
        \centering
        \includegraphics[width=\textwidth]{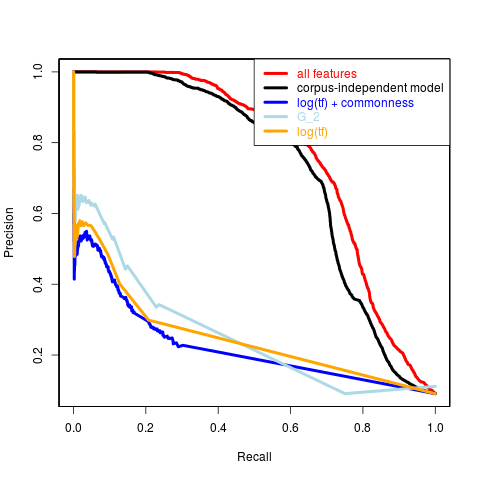}
        \caption{After replication}
        \label{fig:adding_positional_features}
    \end{subfigure}
    \caption{Adding positional features on top of all other features (best viewed in color).}
    \label{fig:adding_positional}
\end{figure}

\begin{figure}
    \centering
    \begin{subfigure}{0.4\textwidth}
        \centering
        \includegraphics[width=\textwidth]{adding_term_commonness.png}
        \caption{No binning}
        \label{fig:adding_term_commonness_subfig}
    \end{subfigure}
    \hfill
    \begin{subfigure}{0.4\textwidth}
        \centering
        \includegraphics[width=\textwidth]{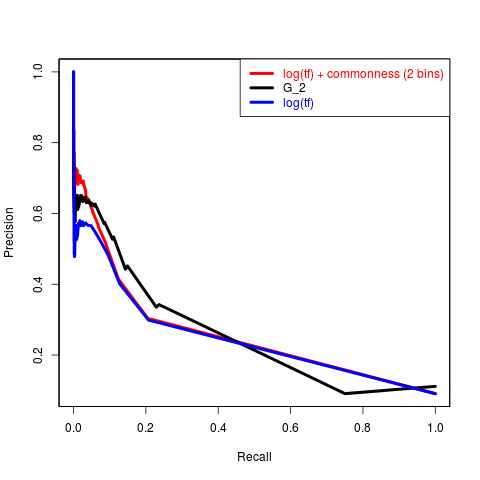}
        \caption{2 bins}
        \label{fig:adding_term_commonness_2_bins}
    \end{subfigure}

    \begin{subfigure}{0.4\textwidth}
        \centering
        \includegraphics[width=\textwidth]{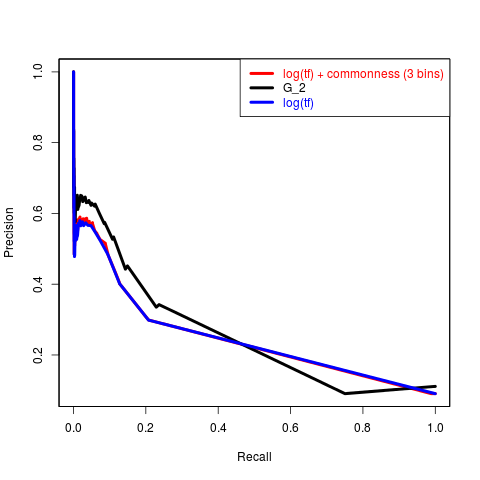}
        \caption{3 bins}
        \label{fig:adding_term_commonness_3_bins}
    \end{subfigure}
    \hfill
    \begin{subfigure}{0.4\textwidth}
        \centering
        \includegraphics[width=\textwidth]{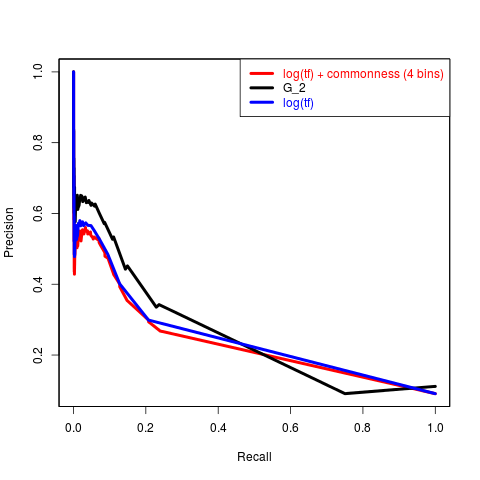}
        \caption{4 bins}
        \label{fig:adding_term_commonness_4_bins}
    \end{subfigure}

    \begin{subfigure}{0.4\textwidth}
        \centering
        \includegraphics[width=\textwidth]{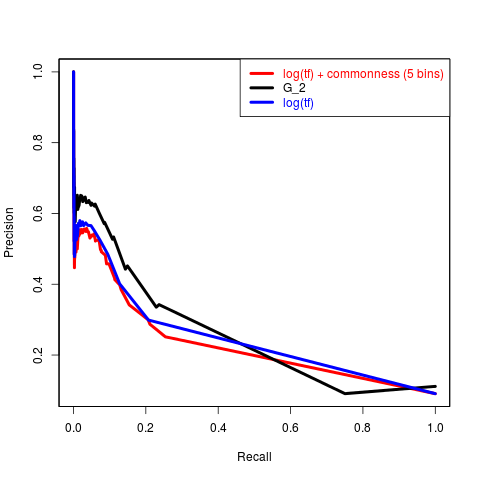}
        \caption{5 bins}
        \label{fig:adding_term_commonness_5_bins}
    \end{subfigure}
    \hfill
    \begin{subfigure}{0.4\textwidth}
        \centering
        \includegraphics[width=\textwidth]{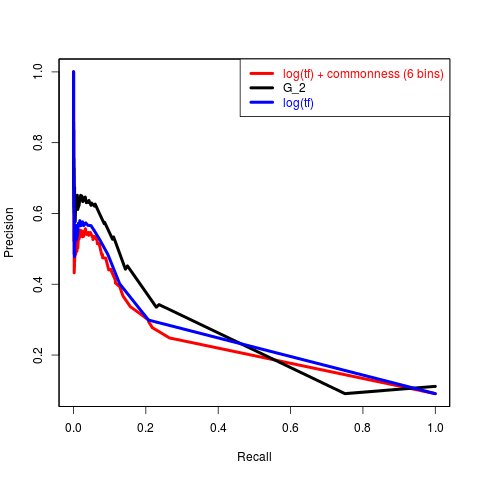}
        \caption{6 bins}
        \label{fig:adding_term_commonness_6_bins}
    \end{subfigure}
    \caption{Binning commonness (best viewed in color).}
    \label{fig:commonness_bins_1}
\end{figure}

\begin{figure}
    \centering
    \begin{subfigure}{0.4\textwidth}
        \centering
        \includegraphics[width=\textwidth]{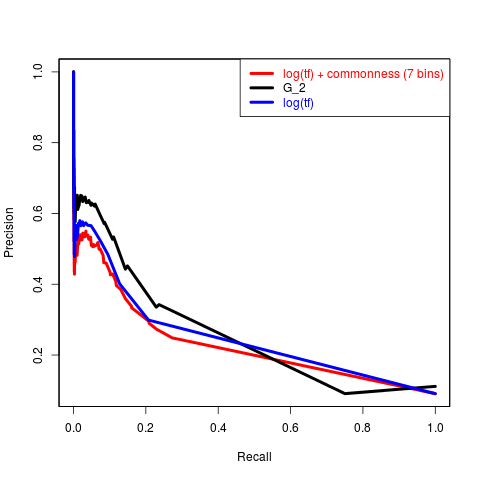}
        \caption{7 bins}
        \label{fig:adding_term_commonness_7_bins}
    \end{subfigure}
    \hfill
    \begin{subfigure}{0.4\textwidth}
        \centering
        \includegraphics[width=\textwidth]{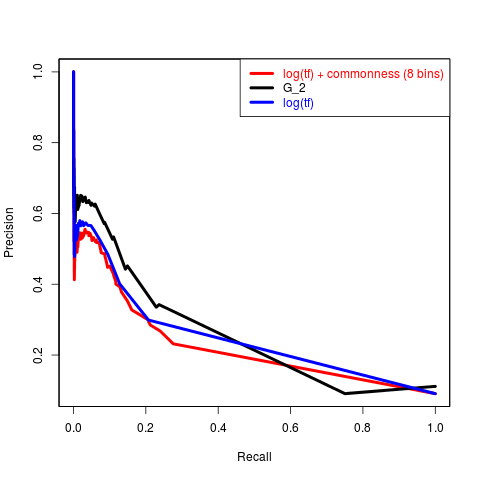}
        \caption{8 bins}
        \label{fig:adding_term_commonness_8_bins}
    \end{subfigure}

    \begin{subfigure}{0.4\textwidth}
        \centering
        \includegraphics[width=\textwidth]{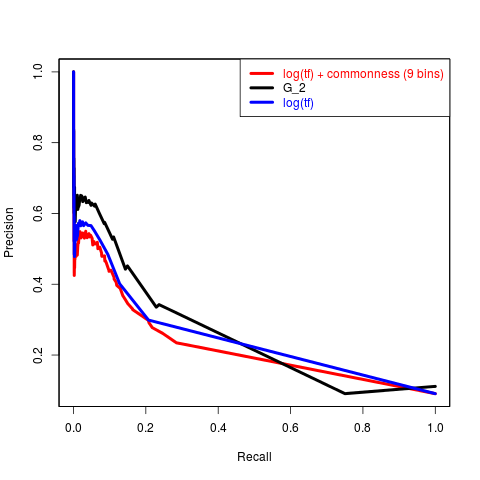}
        \caption{9 bins}
        \label{fig:adding_term_commonness_9_bins}
    \end{subfigure}
    \hfill
    \begin{subfigure}{0.4\textwidth}
        \centering
        \includegraphics[width=\textwidth]{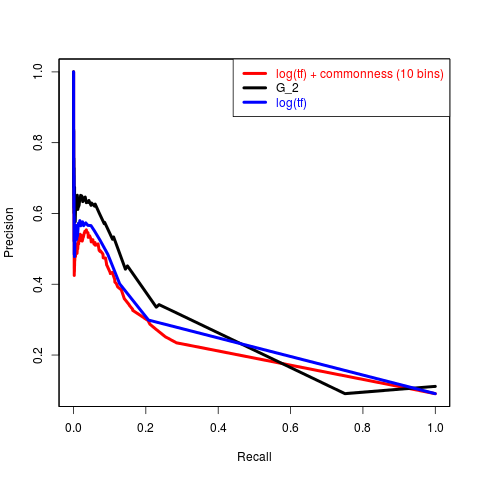}
        \caption{10 bins}
        \label{fig:adding_term_commonness_10_bins}
    \end{subfigure}

    \begin{subfigure}{0.4\textwidth}
        \centering
        \includegraphics[width=\textwidth]{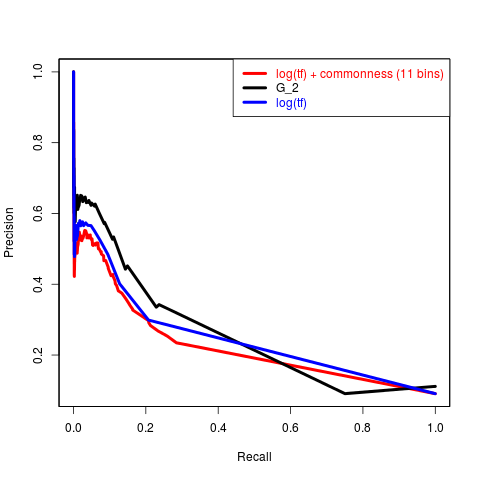}
        \caption{11 bins}
        \label{fig:adding_term_commonness_11_bins}
    \end{subfigure}
    \hfill
    \begin{subfigure}{0.4\textwidth}
        \centering
        \includegraphics[width=\textwidth]{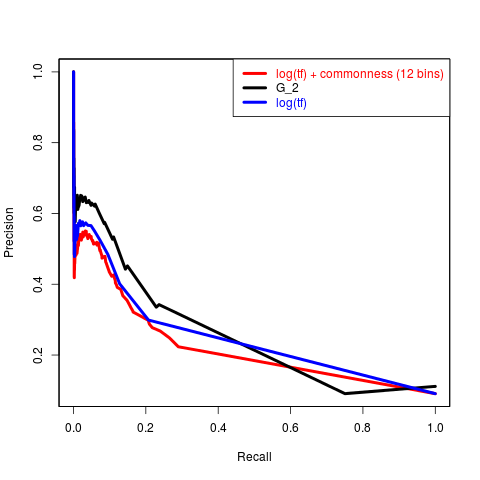}
        \caption{12 bins}
        \label{fig:adding_term_commonness_12_bins}
    \end{subfigure}
    \caption{Binning commonness (best viewed in color).}
    \label{fig:commonness_bins_2}
\end{figure}

\begin{figure}
    \centering
    \begin{subfigure}{0.4\textwidth}
        \centering
        \includegraphics[width=\textwidth]{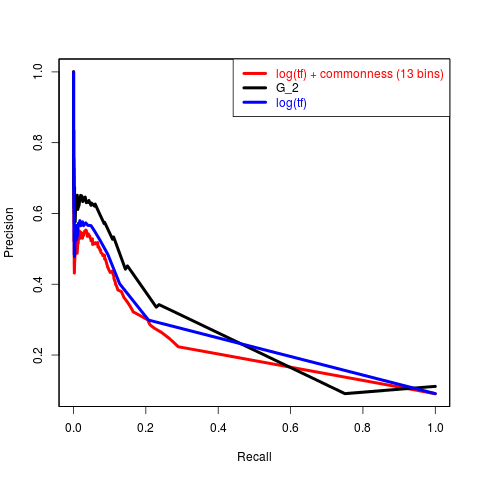}
        \caption{13 bins}
        \label{fig:adding_term_commonness_13_bins}
    \end{subfigure}
    \hfill
    \begin{subfigure}{0.4\textwidth}
        \centering
        \includegraphics[width=\textwidth]{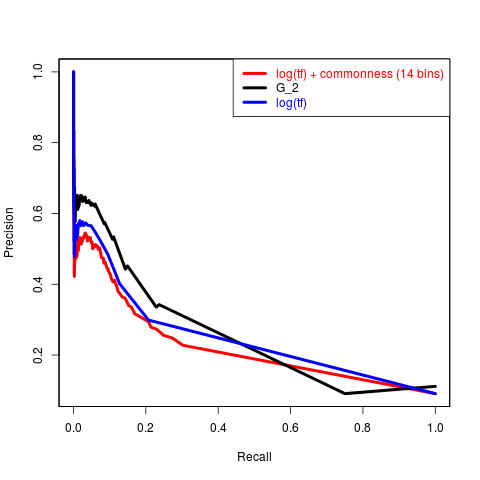}
        \caption{14 bins}
        \label{fig:adding_term_commonness_14_bins}
    \end{subfigure}

    \begin{subfigure}{0.4\textwidth}
        \centering
        \includegraphics[width=\textwidth]{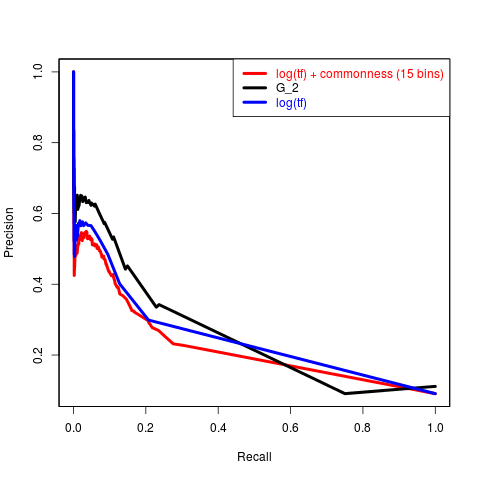}
        \caption{15 bins}
        \label{fig:adding_term_commonness_15_bins}
    \end{subfigure}
    \hfill
    \begin{subfigure}{0.4\textwidth}
        \centering
        \includegraphics[width=\textwidth]{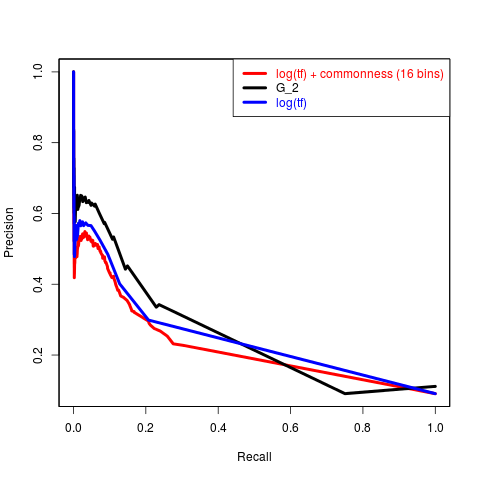}
        \caption{16 bins}
        \label{fig:adding_term_commonness_16_bins}
    \end{subfigure}

    \begin{subfigure}{0.4\textwidth}
        \centering
        \includegraphics[width=\textwidth]{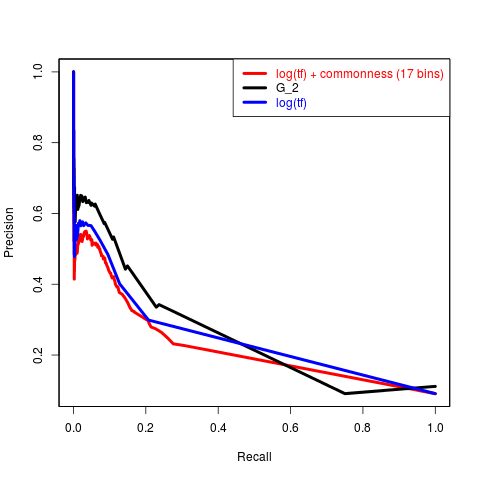}
        \caption{17 bins}
        \label{fig:adding_term_commonness_17_bins}
    \end{subfigure}
    \hfill
    \begin{subfigure}{0.4\textwidth}
        \centering
        \includegraphics[width=\textwidth]{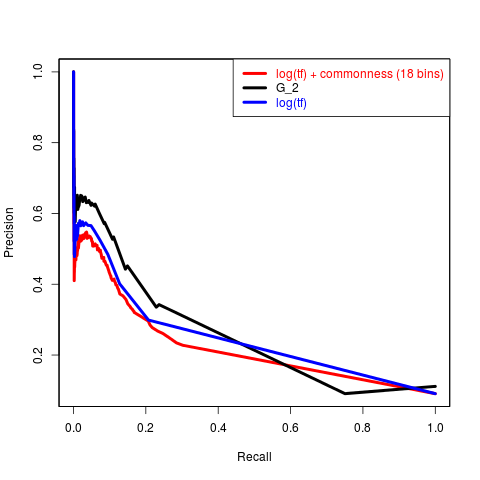}
        \caption{18 bins}
        \label{fig:adding_term_commonness_18_bins}
    \end{subfigure}
    \caption{Binning commonness (best viewed in color).}
    \label{fig:commonness_bins_3}
\end{figure}

\begin{figure}
    \centering
    \begin{subfigure}{0.4\textwidth}
        \centering
        \includegraphics[width=\textwidth]{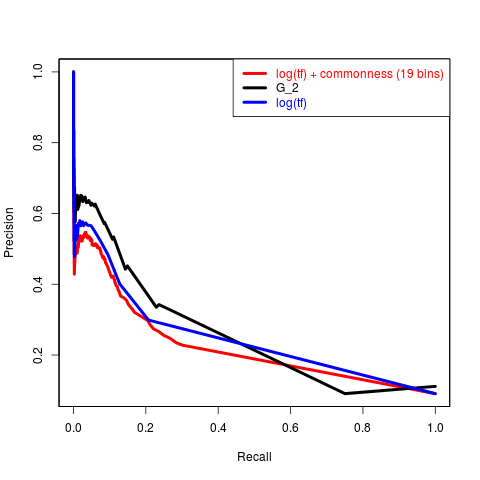}
        \caption{19 bins}
        \label{fig:adding_term_commonness_19_bins}
    \end{subfigure}
    \hfill
    \begin{subfigure}{0.4\textwidth}
        \centering
        \includegraphics[width=\textwidth]{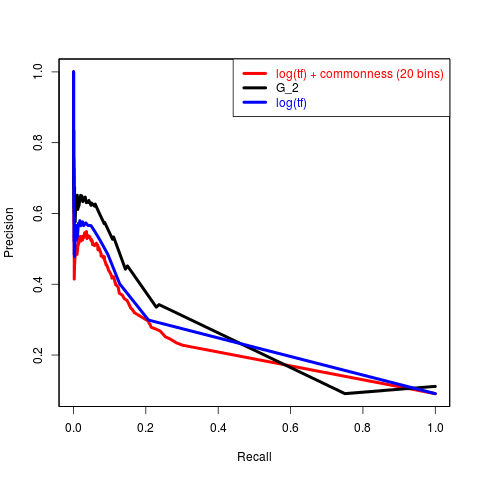}
        \caption{20 bins}
        \label{fig:adding_term_commonness_20_bins}
    \end{subfigure}
    \caption{Binning commonness (best viewed in color).}
    \label{fig:commonness_bins_4}
\end{figure}

Lastly, we added positional features on top of all other features (the best-performing model) to give a final boost in performance. Results look similar to the original. Compare Figure \ref{fig:adding_positional_features} with Figure \ref{fig:4}. \emph{All features} outperform \emph{corpus-independent model} (by a thin margin), which in turn outperforms $G^2$, log(tf), and log(tf) + commonness combination. Note that the corpus-independent model only includes log(tf), positional, and part-of-speech features, and is therefore independent of the \emph{background corpus}.

The authors mentioned \emph{binning} commonness values to see if it has an effect on performance. They did not report precision-recall curves on this, but we went ahead and carried out the experiment nonetheless. Note that our background corpus (Inspec) is way smaller than Web1T, so we are unlikely to see a lot of performance improvement from commonness binning, simply because we do not have a large range of commonness values (like Web1T would have, for example) to bin over. Indeed, we did not see any performance improvement from commonness binning. In fact, performance degraded when the number of bins increased beyond a point. The relevant figures are shown in Figure \ref{fig:commonness_bins_1} through Figure \ref{fig:commonness_bins_4}. We start with the ``adding term commonness'' model of Figure \ref{fig:adding_term_commonness} (no binning; repeated in Figure \ref{fig:adding_term_commonness_subfig}), than gradually increase the number of bins to 2, 3, 4, \ldots, 20. As we can see, performance increases a little when the number of bins is 2, then starts to degrade, and finally reaches a stable (but degraded) configuration when the number of bins is around 6-7. In short, binning did not help much in our case.

\section{Conclusion}
\label{sec:conclusion}

Keyword extraction is a very broad and burgeoning field in Natural Language Processing, and while we do have several references in our bibliography that point to the relevant literature in question, we could by no means justify the exposition of such a vast literature within the scope of this report. There does seem to be very little work done in the specific area of \emph{descriptive keyphrase extraction}, which is the problem tackled by the authors of this paper. There have been some work in information visualization that cited this paper, but none in particular looked at \emph{descriptive keywords}. Our contributions include a re-evaluation and re-assessment of three key hypotheses presented in the paper -- one related to the exploratory analysis of human-generated keyphrases, and two related to the types of features that are good at identifying descriptive keyphrases. Along the way, we also pointed out several differences between our replication study and the original, uncovered some potential limitations of the latter (esp. with respect to \emph{web commonness} histograms), and extended the study by investigating the impact of binning \emph{corpus commonness}. We believe that our efforts will result in the elicitation of interest in this nascent body of work that attempts to integrate two apparently disparate lines of research -- Natural Language Processing and Human-Computer Interaction -- into a unified whole.

\bibliographystyle{plain}
\bibliography{for_report}

\end{document}